\pdfoutput=1

\documentclass[11pt]{article}

\usepackage[preprint]{acl}

\usepackage{times}
\usepackage{latexsym}

\usepackage[T1]{fontenc}

\usepackage[utf8]{inputenc}

\usepackage{microtype}

\usepackage{inconsolata}

\usepackage{graphicx}
\usepackage{hyperref}
\usepackage{float}

\usepackage{listings}
\usepackage{xcolor}

\lstdefinestyle{mystyle}{
  basicstyle=\ttfamily\footnotesize,
  breaklines=true,
  breakindent=0pt, 
  backgroundcolor=\color{gray!10},
  frame=single,
  tabsize=1,
  columns=fullflexible,
  captionpos=b,
  showstringspaces=false, 
  showtabs=false
}
\title{GETALP@AutoMin 2025: Leveraging RAG to Answer Questions based on Meeting Transcripts}

\author{
 \textbf{Jeongwoo Kang},
 \textbf{Markarit Vartampetian},
 \textbf{Felix Herron},
 \textbf{Yongxin Zhou},
 \\
 \textbf{Diandra Fabre},
 \textbf{Gabriela Gonzalez-Saez}
\\
\\
 Univ. Grenoble Alpes, CNRS, Grenoble INP\thanks{Institute of Engineering Univ. Grenoble Alpes}, LIG, 38000 Grenoble, France 
\\
 \small{
   \textbf{Correspondence:} \href{mailto:gabriela-nicole.gonzalez-saez@univ-grenoble-alpes.fr}{firstname.lastname@univ-grenoble-alpes.fr}
 }
}

\begin{document}
\maketitle
\begin{abstract}
This paper documents GETALP's submission to the Third Run of the Automatic Minuting Shared Task at SIGDial 2025. We participated in Task B: question-answering based on meeting transcripts.
Our method is based on a retrieval augmented generation (RAG) system and Abstract Meaning representations (AMR). We propose three systems combining these two approaches.  
Our results show that incorporating AMR leads to high-quality responses for approximately 35\% of the questions and provides notable improvements in answering questions that involve distinguishing between different participants (e.g., who questions).
\end{abstract}

\section{Introduction}

The 2025 edition of the Automatic Minuting (AutoMin) Shared Task introduces, for the first time, a question-answering challenge based on extensive meeting transcripts. 
This task (task B) involves generating accurate answers grounded in long conversational data.

To address this challenge, we propose a retrieval-augmented generation (RAG) approach enriched with Abstract Meaning Representation (AMR). Specifically, we leverage Information Retrieval (IR) techniques to identify and extract relevant passages from large transcripts based on a given question.  Relevant passages are identified using both dense sentence embeddings and synthetic queries generated via the Doc2Query model~\cite{nogueira2019document}. 
To represent the relationships described in the meeting, we include a Knowledge graph from an AMR of the retrieved sentences. These graphs are then translated into natural language descriptions. Finally, we utilize the capabilities of large language models (LLMs) to generate accurate responses using both the user question and the retrieved context.
Our approach thus consists of two main stages: (1) context construction, and (2) answer generation.

To analyze the impact of AMR-based context, we develop and evaluate three system variants:
\begin{enumerate}
\item \textbf{IR-only:} Using only the retrieved sentences (from sentence and Doc2Query representations).
\item \textbf{IR+AMR:} Using both the retrieved sentences and their AMR natural language descriptions.
\item \textbf{AMR-only:} Using only the AMR natural language descriptions of the retrieved sentences.
\end{enumerate}

Finally, we evaluate each variant using the LLM-as-Judge metric~\cite{kim2023prometheus}, and we further conduct a manual evaluation to qualitatively assess the performance of the systems using the same scale.

\section{Related Work}

\subsection{QA based on Meeting Transcripts}
Previous work on question answering from meeting transcripts has explored both extractive and generative approaches. \citet{apel2023meeqanaturalquestionsmeeting} address real questions in meeting dialogues using an extractive model that jointly predicts answers and detects when no answer is present; the authors report moderate performance and note the difficulty of handling ambiguous or unanswered questions. \citet{prasad-etal-2023-meetingqa} use models like Longformer and RoBERTa to extract multi-span answers from full or partial transcripts, but highlight that performance remains well below human level due to the complexity of long, dispersed dialogues. \citet{pan-etal-2024-llmlingua} propose a two-step approach that first compresses transcripts using summarization, then applies QA models to the shortened text; results improve with compression, though performance depends heavily on the quality of the summaries. \citet{golany-etal-2024-efficient} introduce a RAG pipeline where relevant segments are retrieved and used to generate answers; this approach improves handling of dispersed information but can be sensitive to retrieval errors.

\subsection{RAG}

Retrieval-Augmented Generation (RAG) \citep{lewis2020retrieval} is a framework designed to enhance the performance of LLMs by incorporating external knowledge through information retrieval. Instead of relying solely on the model's parametric memory, RAG systems retrieve relevant documents from a collection —such as a database or the Internet— and use these documents as additional context to ground the model's generation. This paradigm has proven effective for injecting up-to-date or domain-specific knowledge into LLMs and improving factual consistency in their outputs. In typical RAG pipelines, user queries are first augmented with retrieved passages, which are then fed into the LLM to generate responses that are both informative and grounded in external sources. A key advantage of RAG is its ability to mitigate the "lost in the middle" phenomenon \citep{liu-etal-2024-lost} —where LLMs overlook relevant content located in the middle of long contexts — by ensuring that only the most relevant content is presented to the model. However, RAG systems also face notable challenges, notably in effectively managing long contexts and multi-document question answering.

\subsection{Meaning representation for question answering}\label{ssec:related_work_AMR}
Previous work leveraged meaning representations for question answering tasks \citep{kapanipathi-etal-2021-leveraging,wang-etal-2023-exploiting}. Meaning representation represents meanings of a text in a structured form such as a graph, tree, or formal logic expressions. Corporating structured information into QA systems provides a few advantages. First, meaning representation reduces ambiguity by explicitly encoding one plausible interpretation among many others. For example, in the following sentence ``Kevin told Tom that \underline{he} broke the glass,'' it is unclear whether `he' refers to Kevin or Tom. This ambiguity can be resolved by explicitly representing its meaning. Second, meaning representation provides information in canonical form regardless of the surface-level variations-especially syntatic ones. For example, ``Mary bought the flower.'' and ``The flower was bought by Mary'' are expressed identically in a meaning representation, thereby reducing the search space in information retrieval systems. Because of these advantages, meaning representation is widely adopted in traditional QA systems.

\begin{figure}[H]
\centering
  \includegraphics[width=.25\textwidth]{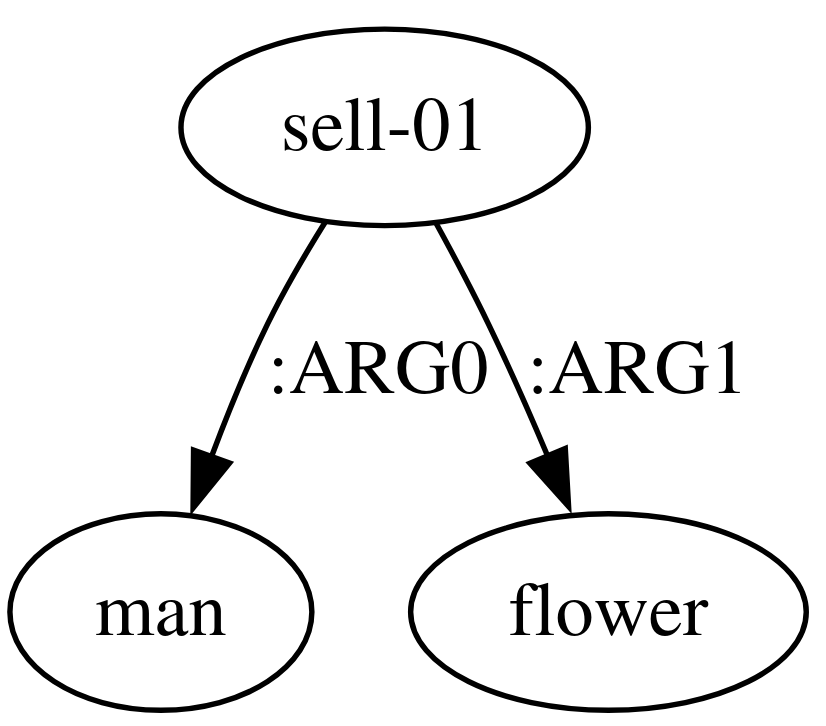}
  \caption{AMR graph for ``A man breaks a window.''}
  \label{fig:amr_graph}
\hfill
\begin{minipage}{.48\textwidth}
  \centering
  \raggedright\vspace{13pt}
  {\texttt{\qquad\qquad(s / sell-01 \\
  \qquad\qquad\qquad:ARG0 (m / man) \\
  \qquad\qquad\qquad:ARG1 (f / flower)) \\}}
  \captionof{figure}{AMR graph linearized in text format.}
  \label{fig:amr_penman}
\end{minipage}
\end{figure}

Among many meaning representation frameworks, Abstract Meaning Representation \citep[AMR]{banarescu-etal-2013-abstract} has gained popularity due to its broad semantic coverage and availability of annotated data. AMR encodes meaning of texts as a rooted, directed and acyclic graph (see Figure \ref{fig:amr_graph}). In AMR graph, the graph nodes are either: Propbank predicate (\textit{e.g.,} sell-01 in Figure \ref{fig:amr_graph}) or English words (\textit{e.g.,} man and flower in Figure \ref{fig:amr_graph}) or AMR-speicifc entities (\textit{e.g.,} date-entity and ordinal-entity). Edges between nodes are labeled to indicate semantic relations between the connected nodes. For example, in Figure \ref{fig:amr_graph}, :ARG0 and :ARG1 respectively indicates that man is the agent of sell-01 and flower is the object of the same predicate. AMR graph can also be serialized in a textual format (see Figure \ref{fig:amr_penman}), which is both human and machine-readable. AMR also uses variables to identify each node, \textit{e.g.,} s, m and f in Figure \ref{fig:amr_penman}. It can also be decomposed into a set of triples that represent the underlying graph structure. 

With ongoing paradigm shift with large langage model, however, the advantage of using AMR as an input for downstream tasks has been questioned. For example, \citet{jin-etal-2024-analyzing} argues that AMR, in its traditional graph form, is not optimal for LLMs, showing that incorporating it offers no improvement across five different NLP subtasks. On the contrary, \citet{zhang-etal-2025-srl} presents evidence supporting the usefulness of AMR when its format is adapted for LLMs. They argue that since LLM is heavily trained with human languages, the structured format of AMR may not align well with their training. To address it, they propose \textit{translating} the graph into a set of textual descriptions by converting each triple of an AMR graph into a natural language sentence. They show that these natural language descriptions of an AMR graph improve the performance of various downstream tasks, both in zero-shot and fine-tuning scenarios. Following their work, we corporate AMR into QA systems while converting its structured form into natural language descriptions.

\section{Methodology}
\subsection{Dataset Description}
We use the two datasets provided for AutoMin TaskB: the ELITR Minuting Corpus and the ELITR-Bench Dataset. 

ELITR Minuting Corpus~\citep{nedoluzhko-etal-2022-elitr} consists of transcripts of meetings in Czech and English. On average, each transcript contains 7,000 words, involves 5.9 speakers, and includes 727 speaker turns. 
ELITR-Bench Dataset~\cite{thonet2024elitr}\footnote{\url{https://github.com/utter-project/ELITR-Bench/tree/main}} contains questions to be answered using the English transcripts from the ELITR Minuting Corpus, splitted in two corpus: Dev and Test. In total, the \textit{Dev} split comprises 10 meetings with 141 questions and is used for model validation prior to submission. The \textit{Test} split, used for the final evaluation in the shared task, includes 8 meetings with 130 questions.
While only the English transcripts are used for the task, the questions are in English (monolingual setting) or in Czech (cross-lingual setting).

\subsection{RAG System Overview} 

Our system follows a two-stage RAG architecture (1) context construction, and (2) Answer generation. 
Given an input Question, denoted as $Q$, and a meeting transcript, denoted as $DOC$, the system produces an answer $A$ that is based on the content of the transcript. 

In the context construction stage, we apply information retrieval (IR) techniques to identify and extract relevant passages from the transcript $DOC$, based on the input question $Q$. We denote this context as the relevant context $C_r$. Using $C_r$ we construct a second context using AMR which is finally translated in natural language, we denote this context as $C_{amr}$. 

In the Answer generation stage, an LLM reads the context $C$ and the query $Q$ to generate the final answer $A$ following a specific prompting strategy.

\paragraph{Context $C_r$ construction: IR}\label{par:info_retrieval}

To construct the relevant context $C_r$, we implement an information retrieval setup that combines two complementary strategies: dense sentence embeddings and Doc2Query-based document expansion. Each sentence in the transcript is not only encoded as a dense vector but also represented by a set of synthetic queries generated using the Doc2Query model~\citep{nogueira2019document}. We index both the sentence embeddings and the synthetic queries using FAISS~\citep{douze2025faisslibrary} for efficient similarity search.

At retrieval time, given a question $Q$, we retrieve (1) the most similar sentences based on the dense embedding similarity to $Q$, and (2) the sentences whose generated queries are most similar to $Q$ in the Doc2Query index. The union of these results forms the initial set of relevant sentences.
To improve coherence, we expand each selected sentence with its immediate context: one preceding and one following sentence from the transcript. We observed that, in some cases, the answer to the question was actually contained in the sentence closest to the most similar one.
The final IR-based context $C_r$ consists of this expanded set of relevant passages, ordered by their original position in the transcript to preserve the sequential structure of the transcript.

\paragraph{Context $C_{amr}$ construction: AMR for QA}
To enrich the retrieved context and improve answer generation, we incorporate AMRs, derived from the selected sentences in $C_r$.
Following the work of \citet{zhang-etal-2025-srl} as described in \ref{ssec:related_work_AMR}, we convert an AMR graph into its natural language descriptions. 
Specifically, we apply this conversion to the context retrieved in the information retrieval step ($C_r$). 
Since the code has yet to be provided by \citet{zhang-etal-2025-srl},\footnote{At the moment of writing, June 2025.} we use our own implementation for this process. We refer the readers to the original article for detailed description and examples. 

Converting AMR into its natural language descriptions consists of 3 steps: 1) Extracting a set of triples from a given AMR graph 2) Translate each triple into a sentence 3) Polish each sentence using LLM. For the first step, we used library \textsc{penman} \citep{goodman-2020-penman}. The second step requires pre-defined rules to translate each semantic role into a sentence, \textit{e.g.,} (John, :ARG0, rob-01) $\rightarrow$ `John is the doer of rob-01 (to engage in or commit robbery)'. This may produce an unnatural text that needs to be polished for natural effect. This step is done in the third step using LLM. Following the original work, we provide some examples for the prompt to polish the text. As a result, for example, `John is the doer of rob-01 (to engage in or commit robbery)' is polished as `John robs something.'

The natural descriptions of AMR graphs form the $C_{amr}$ context, which can be provided either alone or alongside with the original sentences depending on our system variant. This is further detailed in the next section. 

\paragraph{Answer $A$ generation: Prompting LLM}

We use a large language model (LLM) as the backbone of the answer generation component. Given the constructed context $C$, which may include the IR-based context $C_r$, the AMR-derived context $C_{amr}$, or both, and the question $Q$, the LLM generates the final answer $A$.

We experiment with three variants of the input context provided to the LLM:
\begin{enumerate}
\item \textbf{IR-only:} Using only the retrieved sentences based on sentence and Doc2Query representations ($C_r$), along with the question $Q$.
\item \textbf{IR+AMR:} Using both the retrieved sentences ($C_r$) and their AMR-based natural language descriptions ($C_{amr}$), along with the question $Q$.
\item \textbf{AMR-only:} Using only the AMR-based natural language descriptions of the retrieved sentences ($C_{amr}$), along with the question $Q$.
\end{enumerate}

\section{Experiments}

\subsection{Models}

We implement our RAG pipeline using the following components : 

\paragraph{Context Construction: IR} For the sentence-level representation in the IR module, we use the all-MiniLM-L6-v2 sentence embedding model\footnote{https://huggingface.co/sentence-transformers/all-MiniLM-L6-v2}. For Doc2Query, we use the DocTTTTTquery model trained on the MS-MARCO dataset\footnote{https://huggingface.co/castorini/doc2query-t5-base-msmarco}. 

\paragraph{Context Construction: AMR}
For AMR-to-text conversion, we use the \texttt{meta-llama/Llama-3.1-8B-Instruct} model\footnote{\url{https://huggingface.co/meta-llama/Llama-3.1-8B-Instruct}}, prompted with the following instruction:
\begin{lstlisting}[language={}]
You are an AI language assistant. Your job is to improve and rewrite a list of sub-sentences (input_sub_sentences) so they flow naturally and resemble fluent, natural language. Use the input_original_sentence as context to guide your rewrites. Follow the format and style shown in the examples. Only output the final polished sentences. Do not include any explanations.
\end{lstlisting}

\paragraph{Answer Generation} 
We use the same \texttt{meta-llama/Llama-3.1-8B-Instruct} model to generate answers, with the following instruction:
\begin{lstlisting}[language={}] 
You are an AI assistant that answers questions using retrieved meeting information. Provide only the most relevant 1-2 sentence answer extracted directly from the content. Follow these rules: Be extremely concise - just the core fact, Use exact terms/phrases from the retrieved content, and never add analysis, disclaimers or "Based on...". 
\end{lstlisting}

\paragraph{From English to Czech} We control the output language by adding the instruction \texttt{``Answer in Czech.''} to the prompt when needed.

\subsection{Evaluation}
For the evaluation of Task B, predictions are evaluated by the LLM-as-a-judge metric, which uses large language models as automated judges to assess the quality of responses. As used in \cite{thonet-etal-2025-elitr,10.5555/3666122.3668142, chiang-lee-2023-large}, these models will compare the system-generated answers with the human-crafted gold reference answer for each given query. In this experiment, Prometheus model was used as implemented by the authors of \cite{kim2023prometheus}. Prometheus is a 13B open-source language model fine-tuned to serve as an evaluator capable of assessing long-form responses based on user-provided rubrics and reference answers. We follow the Prometheus scale of 0 to 5, where 0 is when an answer is not generated in the intended language, and 5 is when the response to evaluate is essentially equivalent to the reference answer. As the final evaluation of the task is in range 0 to 10, we rescale our scores accordingly.

\section{Results}

In this section, we propose an evaluation of the model based on LLM-as-judges for the Czech dataset only, and both LLM-as-judges and human evaluation for the English dataset. We evaluate the significant difference between the different experiments using a t-test. 

\paragraph{Czech results}

\begin{table}[]
    \centering
    \begin{tabular}{|l|r|}
         \hline
         Model & Mean $\pm$ 1 std \\
         \hline
         GETALP@AutoMin & 5.15 $\pm$ 3.73 \\
         GETALP@AutoMin\_amr & 4.97 $\pm$ 3.77 \\
         GETALP@AutoMin\_amr\_only & 4.31 $\pm$ 3.52\\
         \hline
    \end{tabular}
    \caption{Mean and standard deviation for LLM-as-judge evaluation on Czech Answers.}
    \label{tab:mean_std_cz}
\end{table}

\begin{figure}[h!]
    \centering
    \includegraphics[width=1\linewidth]{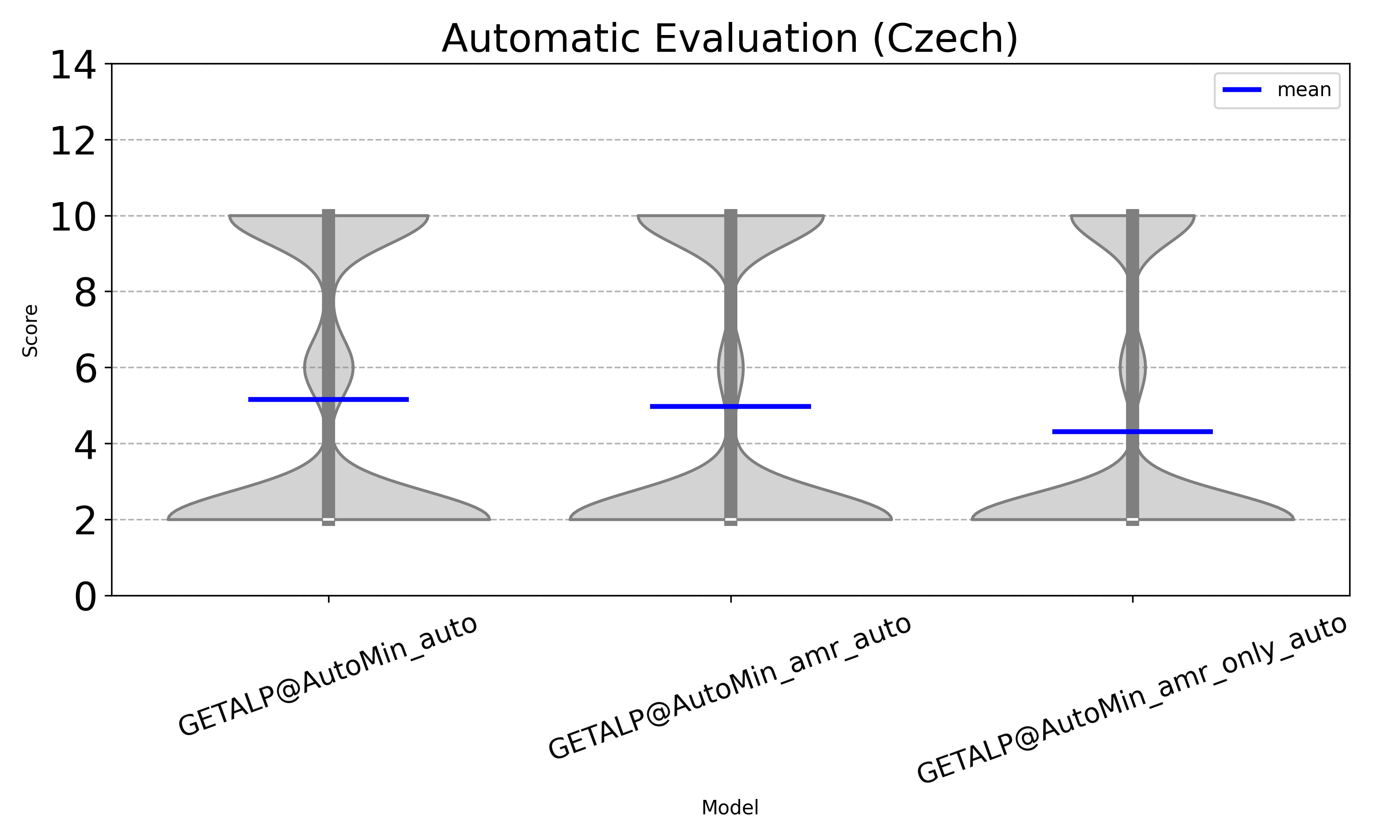}
    \caption{Automatic evaluation using LLM-as-judges. Scores are between 0 and 10. Violin plot with mean distribution as blue line. *** for p$\leq$0.005}
    \label{fig:auto_cz}
\end{figure}

Table~\ref{tab:mean_std_cz} shows the mean and standard deviation from the scores as provided by LLM-as-judges. Figure~\ref{fig:auto_cz} displays violin plots of the score distribution. 
We decided to remove the 0 score, as it was also triggered in cases where the reference answer was in English (e.g., \texttt{[ORGANIZATION1]}).
 
No significant difference was observed between the results obtained for each of our three proposed architectures.

\paragraph{English results}

\begin{table}[]
    \centering
    \begin{tabular}{|l|r|}
         \hline
         Model & Mean $\pm$ 1 std \\
         \hline
         \multicolumn{2}{c}{\textbf{LLM-as-Judge}} \\
         \hline
         GETALP@AutoMin & 4.09 $\pm$ 3.16 \\
         GETALP@AutoMin\_amr & 3.35 $\pm$ 2.54 \\
         GETALP@AutoMin\_amr\_only & 2.46 $\pm$ 1.75 \\
         \hline
         \multicolumn{2}{c}{\textbf{Human evaluation}} \\
         \hline
         GETALP@AutoMin & 5.65 $\pm$ 3.06 \\
         GETALP@AutoMin\_amr & 5.55 $\pm$ 2.95 \\
         GETALP@AutoMin\_amr\_only & 3.94 $\pm$ 2.69\\
         \hline
    \end{tabular}
    \caption{Mean and standard deviation for LLM-as-judge and Human evaluation on English Answers.}
    \label{tab:mean_std_en}
\end{table}

\begin{figure}[h]
    \centering
    \includegraphics[width=1\linewidth]{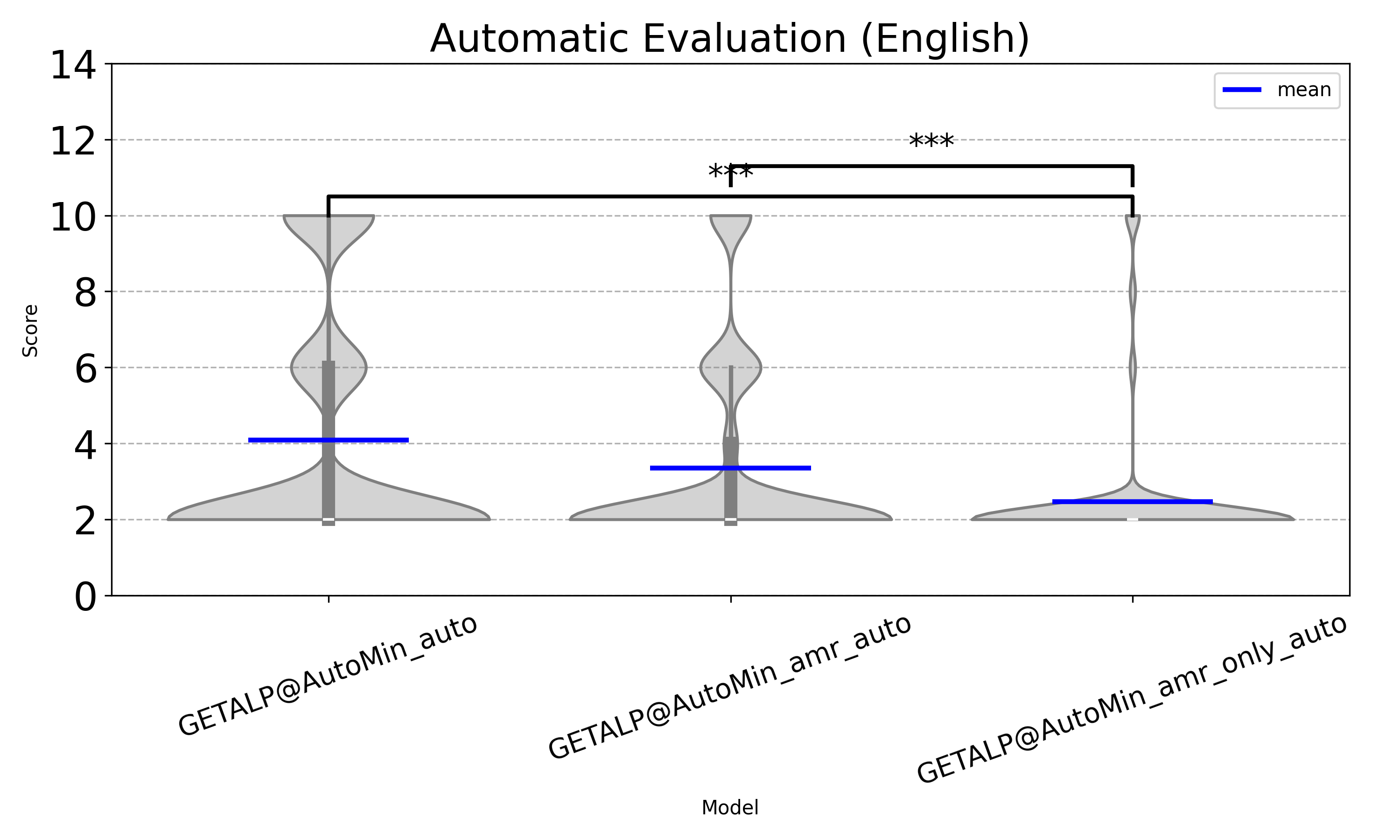}
    \caption{Automatic evaluation using LLM-as-judges. Scores are between 2 and 10. Violin plot with distribution mean as blue line. *** for p$\leq$0.005}
    \label{fig:auto_en}
\end{figure}

\begin{figure}[h]
    \centering
    \includegraphics[width=1\linewidth]{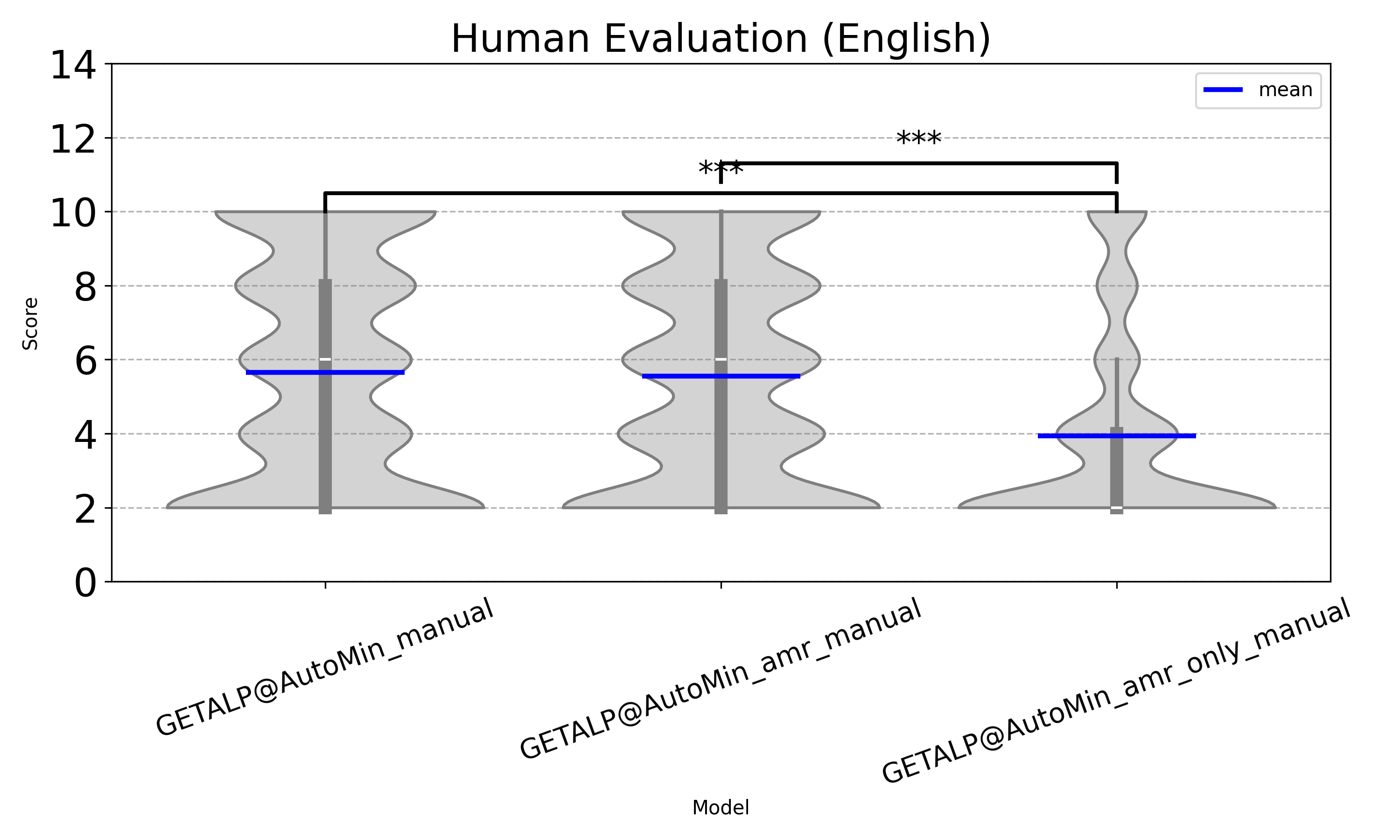}
    \caption{Human evaluation. Scores are between 2 and 10. Violin plot with distribution mean as blue line. *** for p$\leq$0.005}
    \label{fig:manual_en}
\end{figure}

\begin{figure}[h]
    \centering
    \includegraphics[width=1\linewidth]{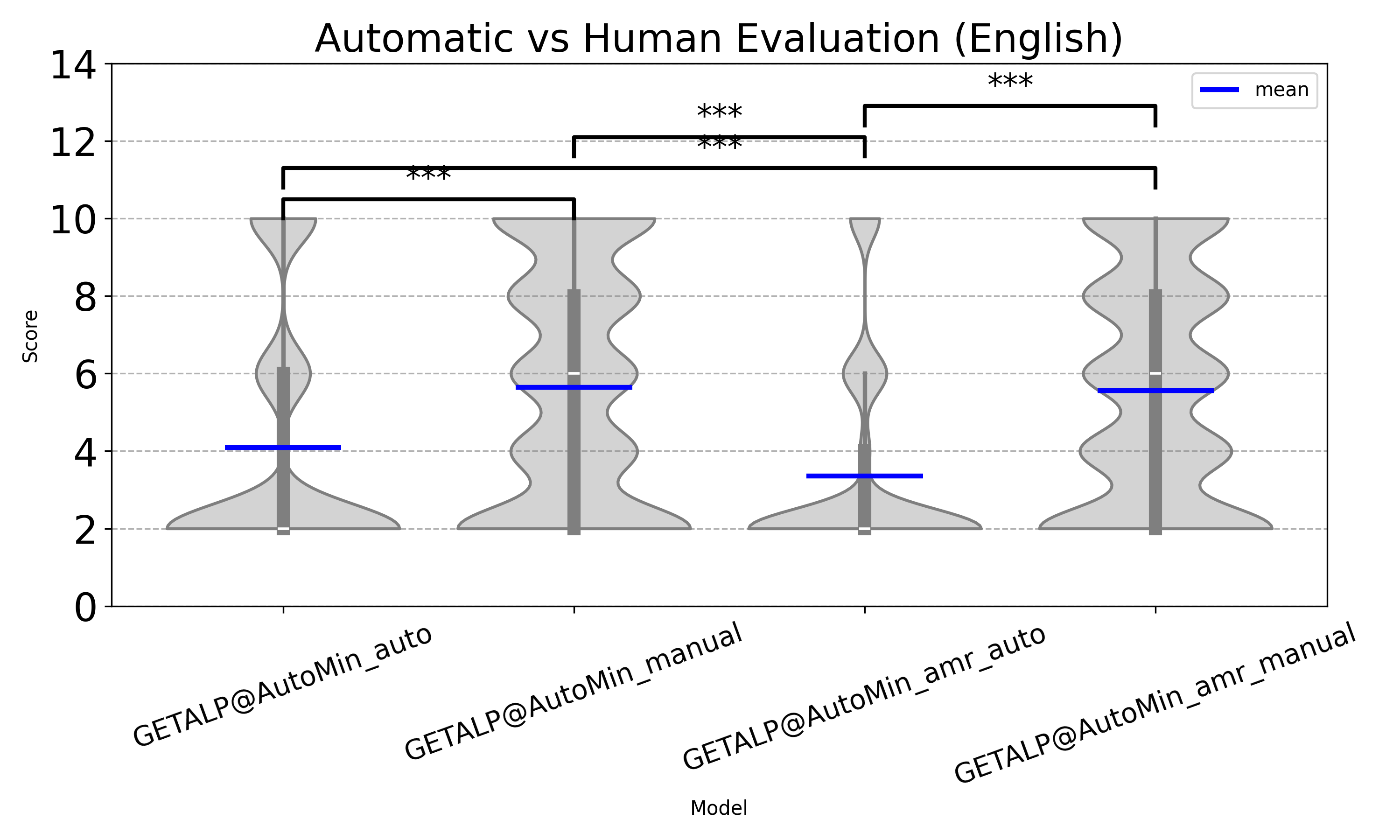}
    \caption{Human versus automatic evaluation. Scores are between 2 and 10. Violin plot with distribution mean as blue line. *** for p$\leq$0.005}
    \label{fig:manual_auto_en}
\end{figure}

We used different human annotators to manually evaluate the performance of each of our three configurations. Each annotator evaluated only a part of the dataset, and there was no cross-over between annotators. Table~\ref{tab:mean_std_en} shows the mean and standard deviation of the scores provided by both LLM and Human evaluators. Human scores are higher than automatic LLM-as-judges scores. We can observe from Figure~\ref{fig:auto_en} and Figure~\ref{fig:manual_en} that the evaluation is consistent between humans and LLMs. In both cases, GETALP@Automin and GETALP@AutoMin\_amr obtain higher scores than GETALP@Automin\_amr\_only, and no significant difference is observed between GETALP@Automin and GETALP@Automin\_amr. When comparing both automatic and manual scores for the two best configurations, as displayed in Figure~\ref{fig:manual_auto_en}, we could not identify a model that outperforms the other.

\begin{table*}[h]
    \centering
    \begin{tabular}{|c|c|}
    \hline
    Ground Truth & Ours \\
    \hline
    \multicolumn{2}{|c|}{\textbf{Who is leaving the project?}} \\
    \hline
    [PERSON10] & [PERSON10] will be leaving. \\
    \hline
    \multicolumn{2}{|c|}{\textbf{What is the application deadline for the PhD program?}} \\
    \hline
    End of the year & The PhD application deadline is the end of the year. \\
    \hline
    \multicolumn{2}{|c|}{\textbf{Who had to pause the meeting for a few minutes?}} \\
    \hline
    [PERSON2] & PERSON2 had to pause the meeting for a few minutes. \\
    \hline
    \end{tabular}
    \caption{Examples where LLM-as-judge gave a score of 2 while evaluators gave a score of either 8 or 10, for both GETALP@Automin and GETALP@Automin\_amr models. The answer displayed comes from one of the two models.}
    \label{tab:manual_good}
\end{table*}

\begin{table*}[h]
    \centering
    \begin{tabular}{|c|c|}
    \hline
    Ground Truth & Ours \\
    \hline
    \multicolumn{2}{|c|}{\textbf{What is [PERSON4] focusing on now?}} \\
    \hline
    Multi-source machine translation & [PERSON4] is putting together and also working on the censorship \\ & component, which is one part of the pipeline after the [PROJECT4]. \\
    \hline
    \multicolumn{2}{|c|}{\textbf{What is the current delay of the language id system?}} \\
    \hline
    3 seconds & The current delay of the language ID system is not explicitly stated.\\
    \hline
    \multicolumn{2}{|c|}{\textbf{How did [PERSON7] qualify the experience of the latest recording session assessment?}} \\
    \hline
    It was qualified as "disastrous". & Unfortunately, the provided information does not contain a direct \\ & quote from [PERSON7] regarding the experience of the latest recording \\ & session assessment.  \\
    \hline
    \end{tabular}
    \caption{Examples where both LLM-as-judge and evaluators gave a score 2, for both GETALP@Automin and GETALP@Automin\_amr models. The answer displayed comes from one of the two models.}
    \label{tab:manual_bad}
\end{table*}

Ground-truth answers provided by the dataset are not always complete sentences, but are often sentence fragments or short pieces of information, as shown in Table~\ref{tab:manual_bad} and Table~\ref{tab:manual_good}. However, given the text generation capabilities of LLMs, we would expect a correct answer to be a full sentence conveying the correct information.
Out of 130 questions, 46 received a human evaluation score of 8 or 10 for GETALP@Automin\_amr, and 49 out of 130 for GETALP@Automin.
The two systems obtained the same score for 91 of the 130 questions, while in 18 questions the AMR-based solution performed better than IR-only. Interestingly, half of these 18 questions correspond to \texttt{WHO} questions. Among the 45 WHO questions in total, AMR achieved the same or a better score in 39 of them.

\section{Conclusion}

Our participation in the AutoMin 2025 Shared Task focused on developing RAG system for question answering over long meeting transcripts. To address the challenges of this task, we combined dense retrieval with Doc2Query-based document expansion and enriched the retrieved content using AMR.
We explored three variants of our system: using only the retrieved passages, combining them with their AMR-based natural language descriptions, and using only the AMR descriptions. 

Our results suggest that AMR contexts can improve the quality of generated answers, particularly for questions involving entity resolution or semantic roles, such as identifying the responsible person for a task or determining who is experiencing an issue (e.g., "Who is experiencing disk space issues?").
Future work includes refining the AMR-to-text generation process, better integrating AMR into context construction, and selectively applying AMR in question types where structured semantic information offers the most benefit.

\section*{Limitations}

Our approach relies on a large language model (Llama 3.1 8B) for both AMR-to-text generation and final answer generation. This significantly increases computational demands and limits the feasibility of our system in resource-constrained environments. To carry out our experiments, we required high-performance GPUs, including an NVIDIA RTX A6000 and NVIDIA H100. Furthermore, although we prompt the model to produce answers in Czech, many of the underlying components, such as sentence embeddings and the Doc2Query model, are primarily trained on English data. This can result in reduced answer quality in non-English outputs and potential inconsistencies in multilingual behavior.

\section*{Acknowledgments}
This work is being carried out as part of the AugmentIA Chair and supported by the Grenoble INP Foundation, thanks to sponsorship from the Artelia Group. This chair also receives state funding managed by the French National Research Agency under France 2030, reference ANR-23-IACL-0006 (MIAI Cluster). 
In addition, this work is supported by the CREMA project (Coreference REsolution into MAchine translation) funded by the French National Research Agency (ANR), contract number ANR-21-CE23-0021-01.

\bibliography{acl_latex}

\end{document}